\documentclass[letterpaper, 10 pt, conference]{./style/ieeeconf}
\usepackage{cite}
\usepackage{amsmath,amssymb,amsfonts}
\usepackage{algorithmic}
\usepackage[pdftex]{graphicx} 
\usepackage{textcomp} 
\usepackage{xcolor}
\usepackage{float}
\usepackage{xspace}

\def\BibTeX{{\rm B\kern-.05em{\sc i\kern-.025em b}\kern-.08em
    T\kern-.1667em\lower.7ex\hbox{E}\kern-.125emX}}

\usepackage{hyperref}
\usepackage{lipsum}
\usepackage{fvextra}
\DefineVerbatimEnvironment{breakverbatim}{Verbatim}{breaklines=true}

\usepackage{cleveref}
\Crefname{equation}{Eq.}{Eqs.} 
\Crefname{figure}{Fig.}{Figs.}

\graphicspath{{images-src/}{images-bin/}} 
\usepackage{comment}
\usepackage{graphicx}
\usepackage{placeins}
\usepackage{cuted}      
\usepackage[caption=false,font=footnotesize]{subfig} 
\usepackage{capt-of} 
\definecolor{dlrprim1}{HTML}{000000} 
\definecolor{dlrprim2}{HTML}{666666} 
\definecolor{dlrprim3}{HTML}{b9cad2}
\definecolor{dlrprim4}{HTML}{ffffff} 

\definecolor{dlrblue1}{HTML}{00658b} 
\definecolor{dlrblue2}{HTML}{3b98cb}
\definecolor{dlrblue3}{HTML}{6cb9dc}
\definecolor{dlrblue4}{HTML}{a7d3ec}
\definecolor{dlrblue5}{HTML}{d1e8fa}

\definecolor{dlryellow1}{HTML}{d2ae3d}  
\definecolor{dlryellow2}{HTML}{f2cd51} 
\definecolor{dlryellow3}{HTML}{f8de53}
\definecolor{dlryellow3}{HTML}{fcea7a}
\definecolor{dlryellow3}{HTML}{fff8be}

\definecolor{dlrgreen1}{HTML}{82a043} 
\definecolor{dlrgreen2}{HTML}{a6bf51}
\definecolor{dlrgreen3}{HTML}{cad55c}
\definecolor{dlrgreen4}{HTML}{d9df78}
\definecolor{dlrgreen5}{HTML}{e6eaaf}

\definecolor{dlrgray1}{HTML}{666666} 
\definecolor{dlrgray2}{HTML}{868585}
\definecolor{dlrgray3}{HTML}{b1b1b1}
\definecolor{dlrgray4}{HTML}{cfcfcf}
\definecolor{dlrgray5}{HTML}{ebebeb}

\usepackage[textsize=scriptsize]{todonotes} 
\newcommand{\quotesasinlatex}[1]{``#1''}
\newcommand{\query}[1]{{``\texttt{\small #1}''}}
\newcommand{\rmcaffiliation}{German Aerospace Center (DLR), Robotics and Mechatronics Center (RMC), M\"unchener Str. 20, 82234 We\ss ling, Germany.}
\newcommand{\tuebaffiliation}{Universität Tübingen, Department of Psychology, Tübingen, Germany.}
\newcommand{\tumaffiliation}{Technical University of Munich, 
Germany.}
\newcommand{\camaffiliation}{University of Cambridge, 
Department of Engineering, Cambridge, UK.}
\newcommand{\iwmaffiliation}{Leibniz-Institut für Wissensmedien (IWM), Tübingen, Germany.}
\IEEEoverridecommandlockouts                              
                                                          
\title{\LARGE \bf
Model Reconciliation through Explainability and Collaborative Recovery in Assistive Robotics
}

\author{Britt Besch$^{1,2}$, Tai Mai$^1$, Jeremias Thun$^1$, Markus Huff$^{3,4}$, Jörn Vogel$^1$,  Freek Stulp$^1$, Samuel Bustamante$^{1,5}$
\thanks{$^1$ \rmcaffiliation}
\thanks{$^2$ \camaffiliation}
\thanks{$^3$ \tuebaffiliation}
\thanks{$^4$ \iwmaffiliation}
\thanks{$^5$ \tumaffiliation}
\thanks{This work was supported by the Bavarian Ministry of Economic Affairs, Regional Development and Energy (StMWi) by means of the projects SMiLE2gether and SMiLE-AI.}
\thanks{Accepted to \textit{IEEE International Conference on Robotics and Automation (ICRA)}, 2026.}
}

\begin{document}

\maketitle
\bstctlcite{IEEEexample:BSTcontrol}


\begin{abstract}
Whenever humans and robots work together, it is essential that unexpected robot behavior can be explained to the user.
Especially in applications such as shared control --- as the name may imply --- the user and the robot must \textit{share} the same model of the objects in the world, and the actions that can be performed on these objects.

In this paper, we achieve this with a so-called model reconciliation framework.
We leverage a Large Language Model to predict and explain the difference between the robot's and the human's mental models, without the need of a formal mental model of the user. 
Furthermore, our framework aims to solve the model divergence after the explanation by allowing the human to correct the robot. 
We provide an implementation in an assistive robotics domain, where we conduct a set of experiments with a real wheelchair-based mobile manipulator and its digital twin. 
\end{abstract}


\section{Introduction}

Whenever humans and robots work together --- such as assistive robots for people with disabilities or cobots in industry --- it is essential that unexpected robot behavior can be explained to the user, as conveying beliefs used in decision-making builds trust~\cite{RN57}.
For instance, a robot may wrongly believe that an appliance is open even though it is closed, and thus fail to provide the user with the support they need. If the robot is able to explain its beliefs about the appliance state, the human can understand the origin of the failed task execution and can correct the robot's knowledge to recover.

To address this challenge, we take inspiration from mental modeling techniques and AI planning, by conceptualizing explanations as a so-called \textit{model reconciliation} process, in which the robot aims to bring the human to a shared understanding of an error situation, comparing the internal models of the robot and the human \cite{RN33, RN39, RN133}. 

In this paper, we consider model reconciliation in the application context of shared control, in particular the control of assistive robots aimed at re-enabling people with motor impairments to perform the activities of daily living in everyday home environments.
We propose a novel framework that leverages a Large Language Model (LLM) workflow \cite{schluntz2024building}:
based on information from the robot's models and the human's queries in natural language, the difference between the robot's and the human's mental model is predicted and explained.  
Our framework allows to explain predicted differences between the user and the robot knowledge bases (for instance a model of the world), but does not require the robot to know the full mental model of the user.

\begin{figure}[t]
	\centering
        \includegraphics[width=0.5\textwidth]{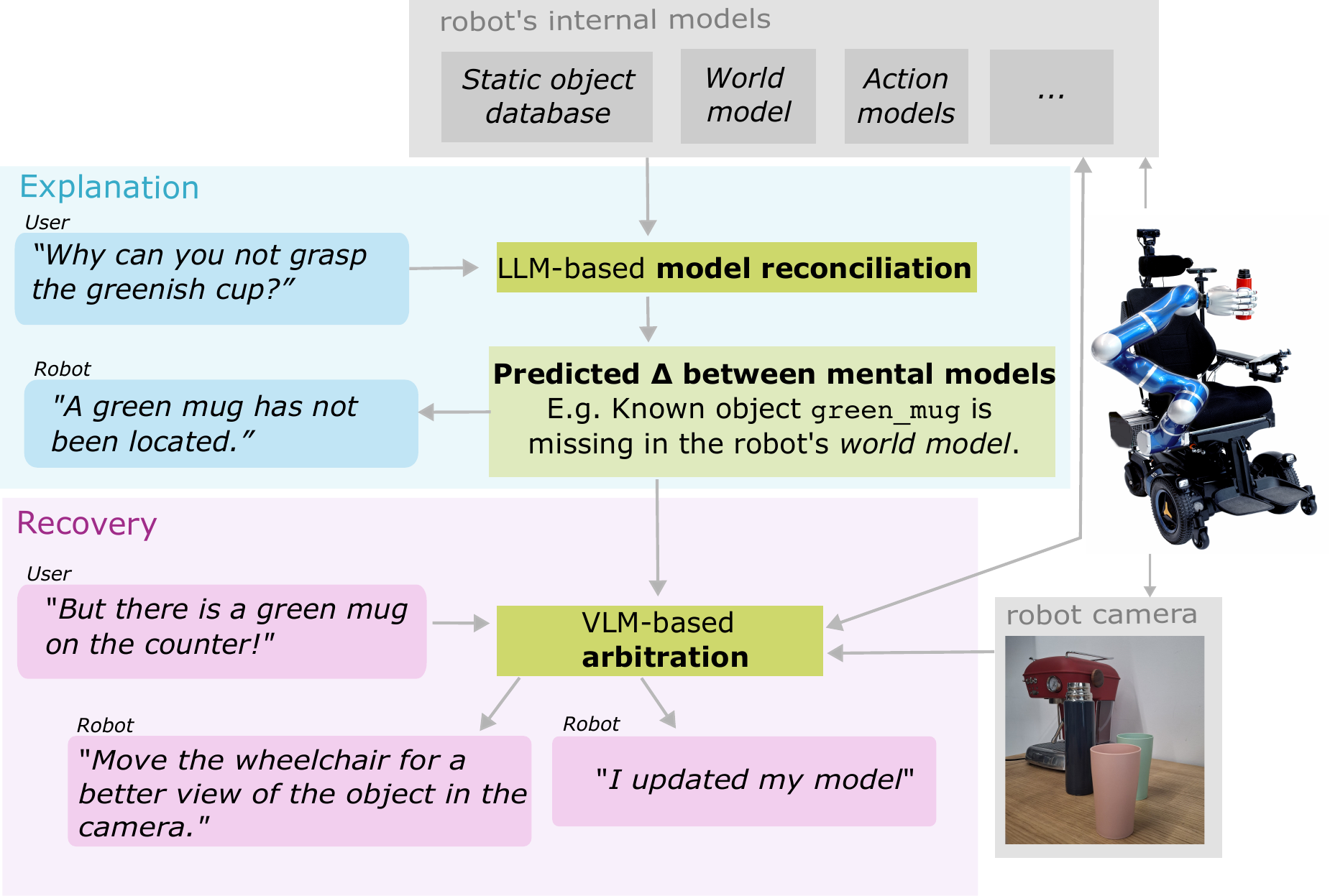}
	\caption{\label{fig:figure_introduction} Overview of the model reconciliation framework including the bidirectional updating process between human and robot, using foundation models.}
\end{figure}

As an example with our wheelchair-based mobile manipulator 
in~\Cref{fig:figure_introduction}, the human inquires why the robot cannot grasp a \textit{greenish cup}. 
The workflow identifies that the human is referring to an object known to the robot as \texttt{green\_mug}. This object is not present in the robot's model of the world, and thus it provides an explanation in natural language.

Furthermore, our framework aims to solve the model divergence after the explanation by allowing the human to correct the robot.
We achieve this with an arbitration module containing a Vision-Language-Model (VLM) that assists the human to recover. Continuing the example in~\Cref{fig:figure_introduction}, the workflow can provide instructions based on the specific robot, such as moving the wheelchair for a better view; or it can overwrite the robot's model by adding the \texttt{green\_mug} directly. 

To summarize, we aim to 1) enable systems to explain their behavior, allowing the human to understand robotic beliefs and 2) create a workflow for recovering from such unexpected situations, particularly when the robot is mistaken, for instance due to perceptual errors. 
The paper makes contributions on three levels: \textbf{firstly}, on a theoretical level by proposing a model reconciliation framework for shared control in assistive robotics, encompassing the robot updating the human and the human updating the robot's model. \textbf{Secondly}, on a methodological level via the usage of foundation model workflows that have been augmented with information retrieved from a set of robot's modules.
This removes the need for the robot to \textit{know} all the human model.
And \textbf{thirdly}, on an experimental level by providing an implementation and a set of experiments in daily living scenarios with our assistive robot, including the real robot and also a digital twin.

\section{Related Work}
\subsection{Background: mental models and theory of mind}
To motivate our framework we draw insights from the social sciences, inspired by Miller \cite{RN1}.
A good explanation needs to be relevant not only to the explainee's questions but also to their mental model. Hence, 
we require not only a model of the explanatory agent and the explainee (we define the construct of a mental model as an internal representation of the current state of affairs~\cite{RN174}), but also the concept of \textit{Theory of Mind}~\cite{RN1}, which describes the ability to attribute mental states, such as beliefs to oneself and others. This ability enables an individual or an agent to predict and interpret the behavior of others based on their presumed mental models. 
Moreover, this also enables humans and agents to understand the so-called \textit{first-order false beliefs} \cite{RN190}. First-order false beliefs involve understanding that another person can hold an incorrect belief about a fact, which is useful in situations where the user and the robot disagree about the world state (e.g. in the green mug example in~\Cref{fig:figure_introduction}). 

\subsection{Explanations as Model Reconciliation}
Agent explanations based on mental models follow a large body of research. 
The AI planning community has coined the term \quotesasinlatex{\textit{model reconciliation}}~\cite{RN133} to describe a set of explainability approaches that combine the robot's model and the robot's model of the human's model as symbolic representations, and provide the differences between them as explanations. 
Several derived works exist, e.g. \cite{RN39,Dung2022,vasileiou2023}.

However, approaches for model reconciliation often focus on explaining the \textit{optimality} of a plan to the human~\cite{vasileiou2023}, and make the assumption that the human's mental model is known to the robot, 
which is not realistic in assistive robotics 
as it is cumbersome for the human to provide the entirety of their mental model. 
Furthermore, many robots with perception systems (e.g. ~\cite{Hagengruber2025}) have dynamic world models that can lead to discrepancies about the world state between the human and the robot models, as in the example in~\Cref{fig:figure_introduction}.
In some cases, the human may not even know that an object or an action is unknown by the robot.  

We take inspiration from previous methods where (1) the robot infers the human model with a dialogue based on a set of rules~\cite{Dung2022}, and (2) explanations reconcile the difference between robot \& human entire knowledge bases, rather than just the optimality of a robot plan~\cite{vasileiou2023}. 
Our approach 
introduces an LLM workflow that exploits its rich semantic understanding of the user's query and robot representations to predict the difference between the human and the robot mental models (such as object databases and world models).
This workflow provides explanations without the need to (1) explicitly represent the human mental models nor (2) limit the explanations to a plan's optimality, for instance by allowing to explain divergent initial world states. 
Our method aligns with a recent workshop paper in which an LLM identifies facts responsible for model misalignment~\cite{moorman2025bidirectional}, and further adds an assistive robotics implementation.

\subsection{Collaborative Failure Recovery}
Note that the methods above provide explanations for robot beliefs. Cases in which the robot's model is mistaken (and therefore not aligned with the ground truth) can be explained but cannot yet be corrected by the human. 
Consequently, this does not necessarily lead to successful failure recovery.

Methods for collaborative failure recovery with a human in the loop mainly focus on the robot detecting failures and asking the human for further assistance for recovery. 
This usually encompasses the human performing manual tasks such as relocating an object for better detection or task execution \cite{RN52, RN50, RN82}. 
\cite{RN80} allows the human to add knowledge via demonstrations for collaborative task decisions and recovery behavior,  
and \cite{RN51} allows the human to intervene via speech with limited vocabulary to guide trajectory assistance for a dressing task. 
More recent work by \cite{RN12} conducts an uncertainty analyses in robot action plans and uses an LLM to actively query the human for additional information collection to avoid task execution if there is incomplete knowledge. 

Yet, approaches that allow for initiation of failure recovery or active intervention by the human are rare. The framework by \cite{RN146} uses an LLM to integrate human initiated corrections into task re-planning and integrates feedback from VLMs for error diagnosis. 
\cite{RN148} further 
distills the human's feedback into a reusable knowledge base to enhance performance in novel settings.
A recent workshop paper \cite{moorman2025bidirectional} proposes a concept with LLMs, where the human is able to provide a verbal interruption to correct the robot, and the LLMs identifies the facts responsible for misalignment through a model reconciliation framework. 
To the best of the author's knowledge, the previous works do not combine an explanatory robotic pipeline with the option for collaborative failure recovery in a model reconciliation framework. 

\section{Model reconciliation framework}
\begin{figure*}[t!] 
	\centering
        \includegraphics[width=1\textwidth]{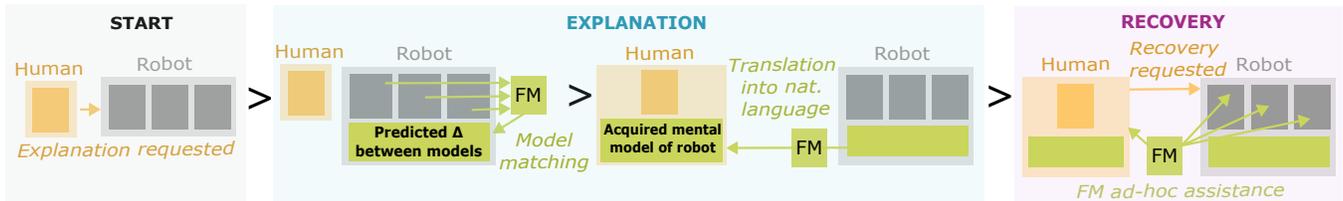}
	\vspace{-1.5 cm} \caption{\label{fig:mental_models_concept} Evolution of the mental models of human and robot during our model reconciliation framework. Yellow boxes represent the user's model. Grey boxes represent the robot's models. See the text for details. \vspace{-0.5 cm} }
\end{figure*}

\subsection{Assumptions}
\textbf{Robot control:} We consider shared control systems where the human and the system jointly manage the robot's control, and focus on an assistive robotics setup (e.g.~\cite{RN191}).
As is common in these setups, we consider robots who schedule one action at the time, guided by the user. 

\textbf{Mental models:} We assume that the human has a mental model of the world as well as of the robot's capabilities. 
Further, we assume that the robot has three models: (1) a database with (static) general knowledge about objects and actions;
(2) a world model, which represents the robot's beliefs about the current state of the world, such as specific object instances and their location (see \cite{Sakagami2023} for examples);
(3) and an action model based on preconditions and effects (e.g.~\cite{Bustamante2022cats}). 
\Cref{implementations} provides details of the implementation we used.
Finally, we also assume that the symbols in those models are semantically grounded in the English language.
As an example, a robot may represent the missing green object in~\Cref{fig:figure_introduction} with a symbol such as \texttt{mug\_green\$2} or \texttt{cup\_green\_id34}, instead of \texttt{object\_blob}.
In this paper, we use the notation \texttt{object\$id} for object instances and \texttt{action\_verb} for actions. 

\subsection{Sources of model divergence}

\newcommand{\Actions}{A}
\newcommand{\Objects}{O}
\newcommand{\Classes}{G}
\newcommand{\Instances}{S}

When the robot and the human are manipulating objects in shared control, there are two main dimensions along which the model of the human and the robot may diverge, related to 
\begin{itemize}
\item knowledge about \textbf{objects} themselves (\Objects) vs about \textbf{actions} that can be performed on objects (\Actions);  
\item \textbf{general} knowledge about objects/actions (\Classes), e.g. ``apples can be picked up'' vs. \textbf{specific} knowledge about objects and the performable action in the \textit{current} world state (\Instances), e.g. ```there is an apple on the table in front of me, but I am currently unable to pick it up''.
\end{itemize}

\newcommand{\Divergence}[2]{\textbf{D}\ensuremath{_{\mbox{\textbf{\footnotesize #1#2}}}}\xspace}
\newcommand{\DivObjCla}{\Divergence{\Classes}{\Objects}}
\newcommand{\DivObjIns}{\Divergence{\Instances}{\Objects}}
\newcommand{\DivActCla}{\Divergence{\Classes}{\Actions}}
\newcommand{\DivActIns}{\Divergence{\Instances}{\Actions}}
\newcommand{\DivFalse}{\textbf{FD}}

From the permutation of these two dimensions, the following four model divergences can be derived:
\begin{itemize}
\item \DivObjCla: The human thinks the robot knows about an object (e.g. a thermos bottle), but it does not (general knowledge about objects).
\item \DivObjIns: The human thinks the robot observes an instance of an object in the current scene (e.g. a specific \texttt{lab\_thermos\$id}), but it does not (specific knowledge about object perception in the current scene). 
\item \DivActCla: The human thinks the robot is able to perform a certain action with an object in general (e.g. pour from bottles), but it is not (general knowledge about actions on objects). 
\item \DivActIns: The human thinks the robot is able to perform a specific action with an object in the current scene (e.g. execute \texttt{pour\_from} from a \texttt{lab\_thermos\$7} into a \texttt{mug\$128}),  but it is currently not due to a missing symbolic precondition (e.g. if it believes \texttt{lab\_thermos\$37} must be grasped first)  (knowledge about preconditions of actions on objects in the current scene).
\end{itemize}

We further add: 
\begin{itemize}
\item \DivFalse: A false divergence, i.e., a human query may imply a divergence, but in reality there is no difference between mental models  (e.g. if the human asks why can the robot not pick up \texttt{lab\_thermos\$37}, but the robot can).
\end{itemize}

\subsection{Model reconciliation}
 A model reconciliation framework for shared control must thus be able to provide explanations so that the human can identify which of the five divergences (\DivObjCla, \DivObjIns, \DivActCla, \DivActIns, \DivFalse) between their model and that of the robot causes the misunderstanding. 
 And it must provide reconciliation methods to help resolve the divergence, if possible. 
We divide this procedure in three stages (start, explanation and recovery) and illustrate the evolution of the mental models in~\Cref{fig:mental_models_concept}.

\subsubsection{Start} 
At the start, both parties are unaware of the other's mental models, and the human asks the robot for an explanation in natural language.
We assume this query is of the form: 
\query{{Why cannot I/cannot you \{action\} \{with an object\}?}}, where we highlight that 
the query does not have to be follow the template exactly, since language models parse it.
Examples: \query{Why can you not grasp the apple?}; \query{Hey robot, I cannot open the drawer, what is the issue?}

\subsubsection{Explanation}
First, we extract the actions and objects (including adjectives such as \quotesasinlatex{greenish}) from the query using an LLM. 
Then, we iteratively search in the robot models to \textit{match} the user query actions (and, if provided, objects and adjectives) in natural language.
Crucial on this step is that LLM's provide a robust semantic interpretation, both of the query and of the robot model contents.

Once a match is found that explains the disagreement, this is communicated to the human in natural language. 
We note in \Cref{fig:mental_models_concept} that this explanation updates the mental model that both parties have of each other, as (1) the human updates their own internal model of the robot; and (2)  
the robot implicitly estimates a human model, as it determines the delta between the models (e.g. the robot learns that the human expects some greenish cup device in \Cref{fig:figure_introduction}).

\begin{figure*}[t]
	\centering
        \includegraphics[width=0.85\textwidth]{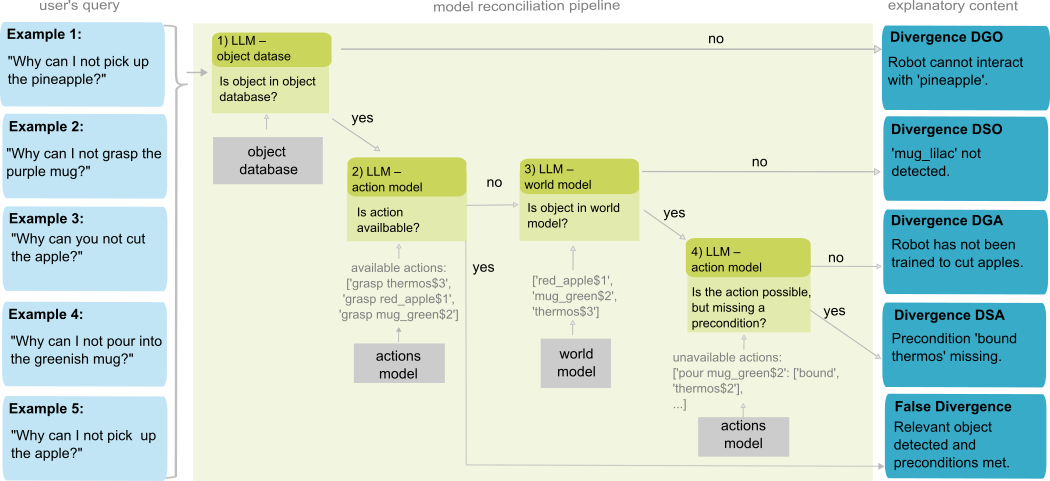}
	\caption{\label{fig:method_explanation} Overview of the explanation generation pipeline. We note that the text in the right column refers to the explanation content and not the final sentence uttered by the LLM.
    See Experiments and attached video for end-to-end examples.}
\end{figure*}

\subsubsection{Recovery procedure}
Provided with an explanation, 
the human decides if they want to accept it and carry on, or provide a natual language rebuttal to the robot instead. 
For instance, if the robot explains a drawer cannot be opened because it believes it is already open, the user can rebut with \query{But the drawer is actually closed!}. 
In this case, the model is either overwritten directly by the estimated difference (in the simplest case), or the robot provides ad hoc methods to recover (e.g. suggesting robot base movements to improve the perspective of perception methods).

We note that if the divergence concerns general knowledge, such as teaching new skills and objects to the robot (divergences \DivObjCla \& \DivActCla), the user may not be able to recover by themselves, as providing this knowledge often requires expert modules. 
In this paper we focus only on recovery of divergences of specific knowledge, namely \DivObjIns \& \DivActIns. 

\subsection{Implementation in assistive robotics}
\label{implementations}

The method is implemented on an wheelchair-based mobile manipulator shown on~\Cref{fig:figure_introduction}, which allows people with severe motor impairments to conduct tasks of the daily living using shared control \cite{RN8}. 

\paragraph{Models}
Our robot represents a static object database (using a framework from~\cite{RN173}) with a list of object classes it knows (e.g. \texttt{green\_cup}), and a world model with a list of object instances it currently locates, using perception (e.g. (e.g. \texttt{[green\_cup\$41, green\_cup\$95]}).

\begin{figure}[b] 
	\centering
	\includegraphics[width=0.95\columnwidth]{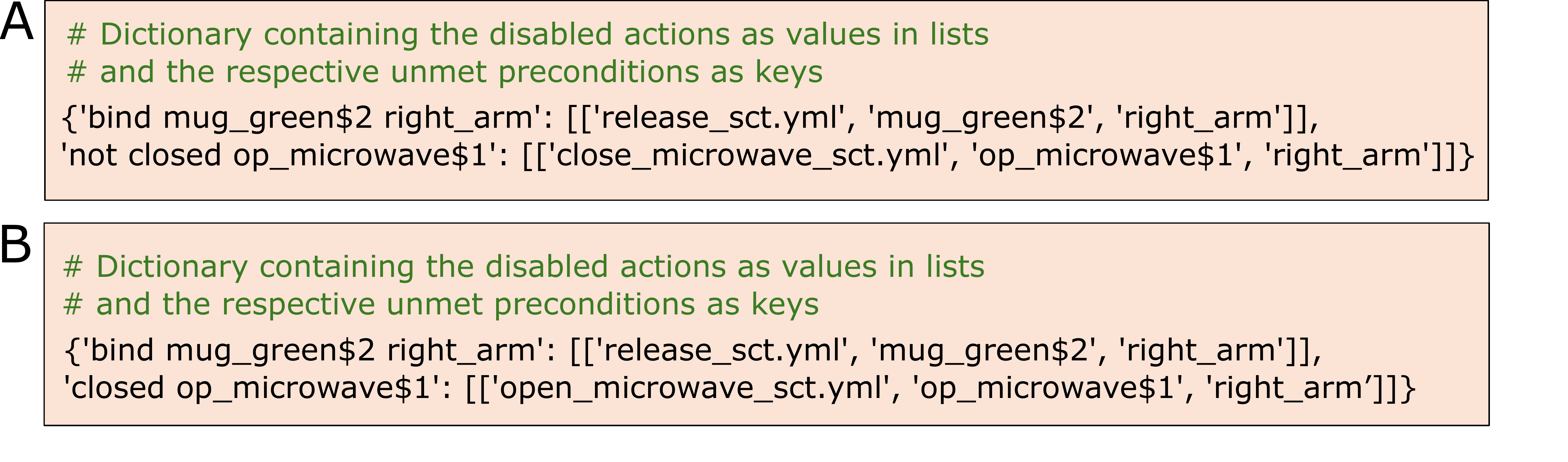}
	\caption{\label{fig:disblededgesdic} Symbolic state dictionaries for a world model with (A) a green mug and a closed microwave; (B) a green mug and an open microwave.}
\end{figure}

For an action model we use a so-called action graph~\cite{Behery2016} that filters actions that are possible for the robot based on the objects in the world model and preconditions in the PDDL language~\cite{RN192}. 
Actions can be either \textit{available} (preconditions are fulfilled, can be executed right away) or \textit{blocked} (preconditions are missing).  
Blocked actions and their associated unmet preconditions are stored in a dictionary (see~\Cref{fig:disblededgesdic}). 

\paragraph{Explanation procedure via model matching}
We implement the explanation procedure by conducting four search steps, where an LLM workflow follows the flowchart in  \Cref{fig:method_explanation}, described below. 
Each leaf node in the flowchart corresponds to a divergence type (\DivObjCla, \DivObjIns, \DivActCla, \DivActIns, \DivFalse), and the figure provides one example for each. 

\begin{figure}[t] 
	\centering
	\includegraphics[width=0.95\columnwidth]{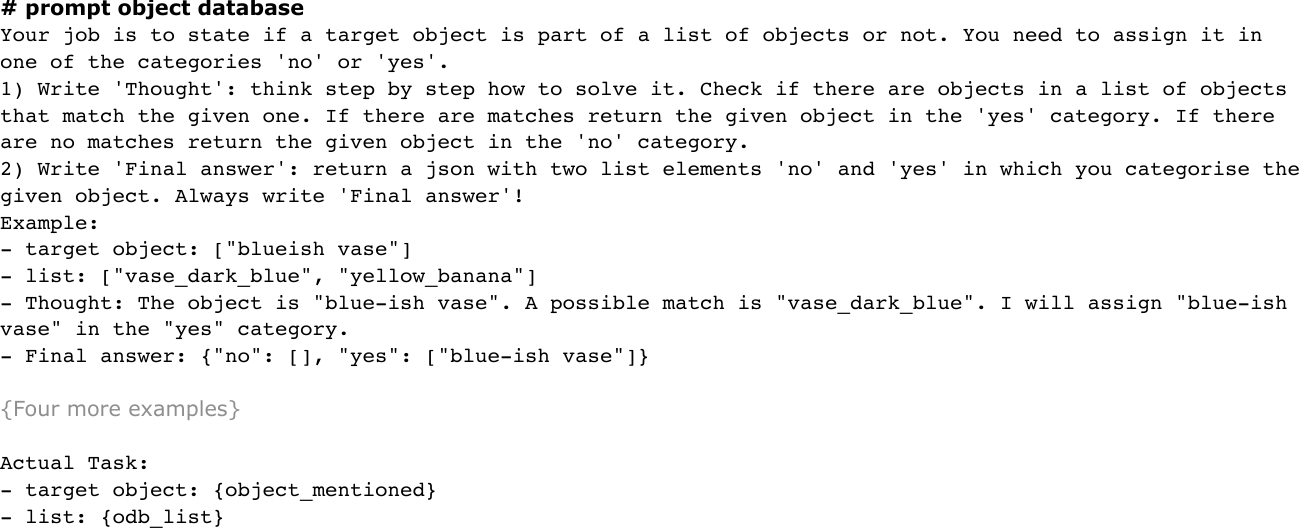}
	\caption{\label{fig:prompt_odb} Example prompt for matching an object with the robot database.}
\end{figure}

\textbf{Step 1 - Query object database (for Divergence \DivObjCla):}
If an object is mentioned in the human’s query, it is first verified that this object 
is part of the robot’s object database. 
The LLM workflow extracts objects from the query and attempts to match their \textit{similarity}
with the objects in the database, see prompt in~\Cref{fig:prompt_odb}, 
and provides an explanation if it fails. 
However, if the object is matched with an object in the database, this object name is saved and used in subsequent search steps. 
\textit{Example}— The human attempts to grasp a pineapple using the robotic arm. However, this is an unknown object to our robot. The method therefore informs the human about the inability to grasp 
the pineapple (see Ex. 1 in \Cref{fig:method_explanation}).  

\begin{figure*}[t]
	\centering
    \includegraphics[width=0.85\textwidth]{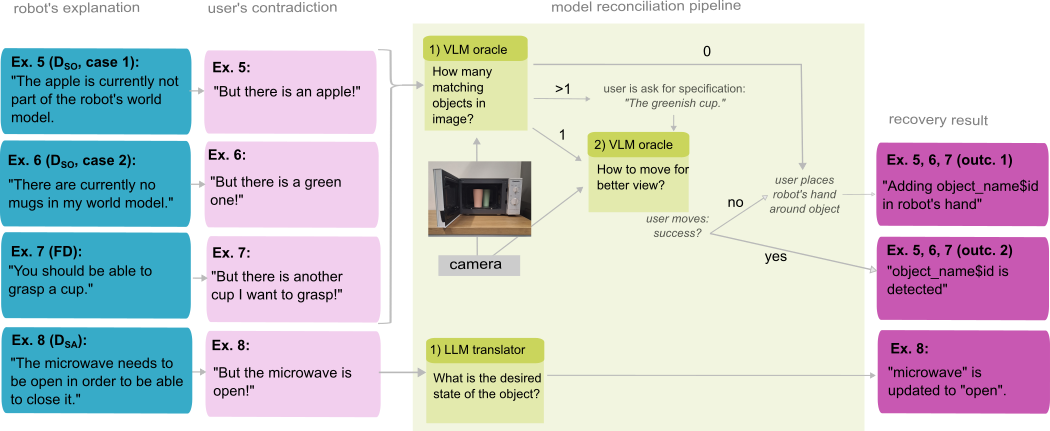}
	\caption{\label{fig:method_recovery} Overview of the recovery workflow.}
\end{figure*}

\textbf{Step 2 - Query action model, first pass (for Divergence \DivFalse):}
It is assessed if the action (with an optional object) in the human's query can actually be executed by the robot, by searching in the action graph's available actions.
This prevents the method from further searching, and requires the human to be more precise if the problem persists. 
\textit{Example}— In Ex. 5 of~\Cref{fig:method_explanation}, the human asks why an apple grasping task is not possible. As the preconditions of these task are met according to the action graph, the robot will simply explain that the grasp is possible after all given the robot models.  
                    
\textbf{Step 3 - Query world model (for Divergence \DivObjIns):} If an object was mentioned in the human's query, the LLM ascertains if an instance of the object name (from Step 1) is part of the robot's world model. 
If it is not, it returns an explanation. 
\textit{Example}— In Ex. 2 in \Cref{fig:method_explanation} the world model contains a \texttt{red\_apple\$1}, \texttt{mug\_green\$2} and \texttt{thermos\$3}, but no object \texttt{mug\_lilac\$id} as identified from the user query about a \quotesasinlatex{purple mug}. 

\textbf{Step 4: Query action model, second pass (for Divergences \DivActCla/\DivActIns):} 
The LLM searches in the blocked action dictionaries from the action graph, aiming to find if the requested action is blocked by a precondition, or not. 
If it is blocked it returns the missing precondition as an explanation (Divergence \DivActIns);
if it is not, we asusme the robot is not trained to execute the said action, and return this as an explanation (Divergence \DivActCla).
\textit{Example for \DivActCla}— The user asks in Ex. 3 about cutting an apple; despite a \texttt{red\_apple\$1} being located, the action cutting is not available nor blocked, so the robot explains the robot has not been trained for it.
\textit{Example for \DivActIns}— The user asks in Ex. 4 about pouring into a green mug. 
Intuitively, grasping a container with liquid is a precondition for this action, but in this case this is missing (e.g., the robot could have failed to detect that a thermos is grasped). As an explanation, the LLM provides the missing precondition from the action graph. 

\paragraph{Recovery procedure}
\label{sec_recovery_procedure}
We implemented a recovery pipeline as a flowchart in~\Cref{fig:method_recovery}.
There are two cases:

\textbf{Case 1: Divergences \DivObjIns \& \DivFalse:}
An object desired by the user is not in the world model (even if the explanation incorrectly assumed a False Divergence, see Ex. 7 in~\Cref{fig:method_recovery}).
We use a VLM as an \quotesasinlatex{oracle} to (1) attempt to match the object in the user rebuttal with the objects in the camera image of the robot, shown in Ex. 6 (successful match) and Ex. 5., (unsuccessful match); and (2) if matched, provide a movement suggestion of the robot wheelchair that would achieve a better object view without occlusions, so that the robot perception algorithms can detect it.
\textit{Outcome 1:}  If there is no match or if the movement of the robot wheelchair do not fix the problem, the LLM adds the missing object to the world model. For convenience, we ask the user to drive the EE towards the object to obtain a pose. 
\textit{Outcome 2:} If the model suggestion works, the robot's perception algorithm will automatically detect the object and add it to the world model. 

\textbf{Case 2: Divergence \DivActIns:}
In case of an unmet precondition, the desired state is collected from the human rebuttal (e.g. the open status of the microwave from \query{But the microwave is open!}). This is matched to a symbolic state (in PDDL), the world model (and the derived action graph) and overwritten accordingly (Ex. 8 in \Cref{fig:method_recovery}).
\paragraph{LLM/VLM}
We used an open-weights local VLM for all steps, \textit{Mistral-Small-3.2-24B-Instruct-2506}\cite{mistral-small-3.2-24b-instruct-2506}.
Prompt engineering techniques such as \textit{Chain of Thought} and \textit{Few Shot Prompting} were applied~\cite{NEURIPS2022_9d560961, gao2021makingpretrainedlanguagemodels}, in order to improve the generation capabilities of the relatively-small model. 

\begin{figure*}[t]
	\centering
    \includegraphics[width=0.95\textwidth]{images-bin/img_robot_expl-svg.pdf}
	\caption{\label{fig:results_expl} First video sequence. Explanation results with a real robot with several objects. The user guides the robot to pick up the mug between frames 4 and 5. We note the world model also represents the user and the end-effector. The full sequence is in the attached video. }
\end{figure*}

\begin{figure*}[t]
	\centering
    \includegraphics[width=0.95\textwidth]{images-bin/img_robot_recov-svg.pdf}
	\caption{\label{fig:results_recov} Second video sequence. Recovery results with a real robot with a thermos and a mug. The full sequence is in the attached video.  }
    \vspace{-0.4cm}
\end{figure*}

\section{Evaluation}

\subsection{Pilot study on real robot}
\textit{Do explanations and recovery work on a realistic setting?}

\textbf{Robot:}
Experiments were conducted on a wheelchair-based mobile manipulator with a lightweight arm and a gripper end-effector, shown in~\Cref{fig:results_expl,fig:results_recov}.
The robot end-effector (EE) was controlled with a 3D joystick in shared control. 
The robot offered to user a set of actions (e.g. a grasp), that could be activated by users by driving the EE nearby the object, and then completed using Shared Control Templates~\cite{RN191}. 
The language model was running on a workstation next to the robot (dual NVIDIA RTX 6000 Ada 48GB GPU), and we interacted with the robot via speech. 

\textbf{Tasks and robot models:}
We considered two sequences where the experimenters commanded the robot to interact with daily living objects, shown in~\Cref{fig:results_expl} and~\Cref{fig:results_recov} respectively, and added to the video submission.
The robot located the objects known to it using a perception pipeline, and updated them on its world and action models. 
Known objects were a thermos bottle, a mug and an apple. 
Unknown objects (i.e., not in the object database) were two toys (pineapple, octopus). 

\textbf{Explanation and recovery:}
On each of the user queries on~\Cref{fig:results_expl,fig:results_recov} the robot did not offer to activate in shared control the tasks that users were asking about (except for divergence FD on Frame 4 of~\Cref{fig:results_expl}).
The first sequence provides examples of the explanation procedure only, highlighting one query of each divergence type.  
The second sequence shows the full explanation and recovery process from Section~\ref{sec_recovery_procedure}, but removing the VLM oracle step with movement suggestions for simplicity of example (featured later on~\Cref{eval_digital_twin}). 
After the recovery interactions on the second sequence, the tasks that were blocked (\quotesasinlatex{grasping the greenish cup}, \quotesasinlatex{grasping the thermos bottle}) were offered again to the user as the issues were solved, and executed in shared control (frames 3 and 6 in~\Cref{fig:results_recov}).

\subsection{Evaluation with a robot simulation}
\label{eval_digital_twin}
We conducted two experiments on a digital twin of the robot, labeled \textbf{Experiment B1} and \textbf{Experiment B2}. 
Exp. B1 tested our method with three independent unit tests:
two units for the generation of correct explanations for world and actions divergences (\DivObjIns, \DivActIns)
, 
and one unit for recovery suggestions in the reconciliation pipeline for a failed object localization (\DivObjIns). 
Exp. B2 revisits the action divergence explanation unit test to evaluate the effect of robot semantic representations. 

\begin{figure}
	\centering
	\includegraphics[width=0.96\columnwidth]{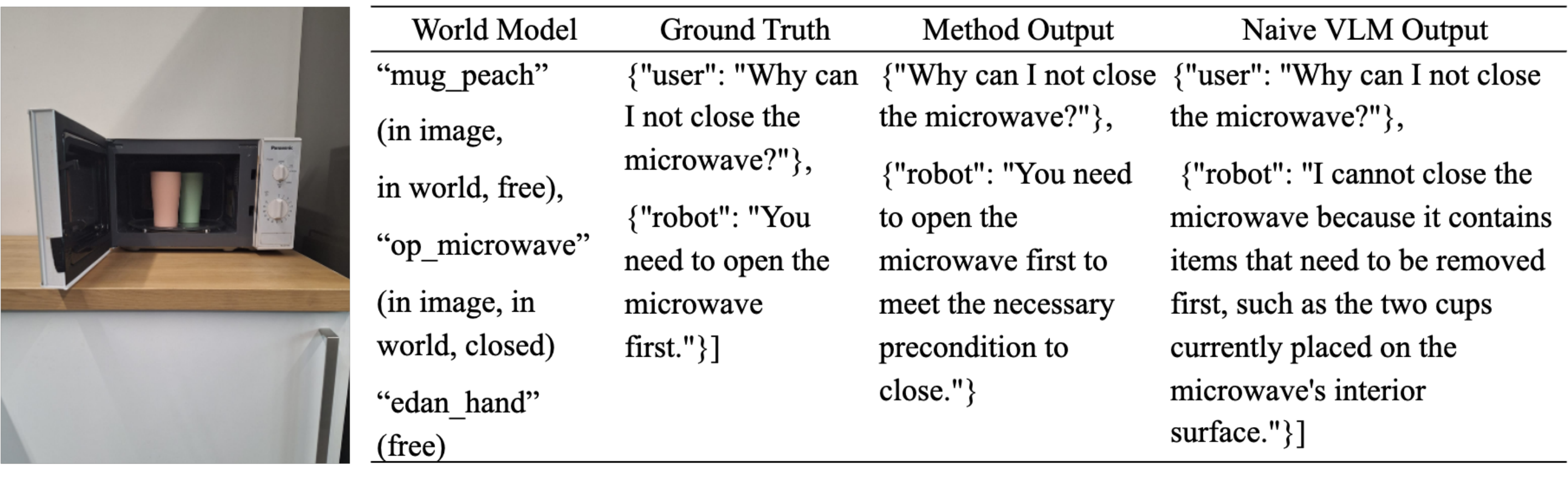}
	\caption{\label{fig:episode} Example episode for the unit testing explanation generation for unmet task precondition}
\end{figure}

\textbf{Dataset:}
A dataset composed of 40 episodes was gathered. Each episode represents one human-robot-interaction for one of the units. There are 13 units for failed object localization, 11 for an unmet predicate and 16 for recovery suggestions. 
An episode consists of an image, a matching world model of the robot, and an initial human's query, plus potential follow-up contradictions or specifications from the human.
The images were partially collected from our robot's camera itself in the kitchen of the lab. Additional image material was gathered in the kitchen and the living room of the lab as well as our office using a smartphone. All images contain at least one object from our robot's object database.
An example for an episode can be seen in \Cref{fig:episode}.
For each episode a ground truth conversation was created. This conversation contains the expected explanations or recovery suggestions based on the given human's query, image, and world model.

\textbf{Answer generation and labeling:} For all experiments (Exp. B1 and B2) we computed corresponding robot explanation or recovery suggestions from a digital twin containing the world and action models of the real robot.
As each episode was then ran three times, we have in total 120 output conversations.

\begin{figure}
	\centering
	\includegraphics[width=1\columnwidth]{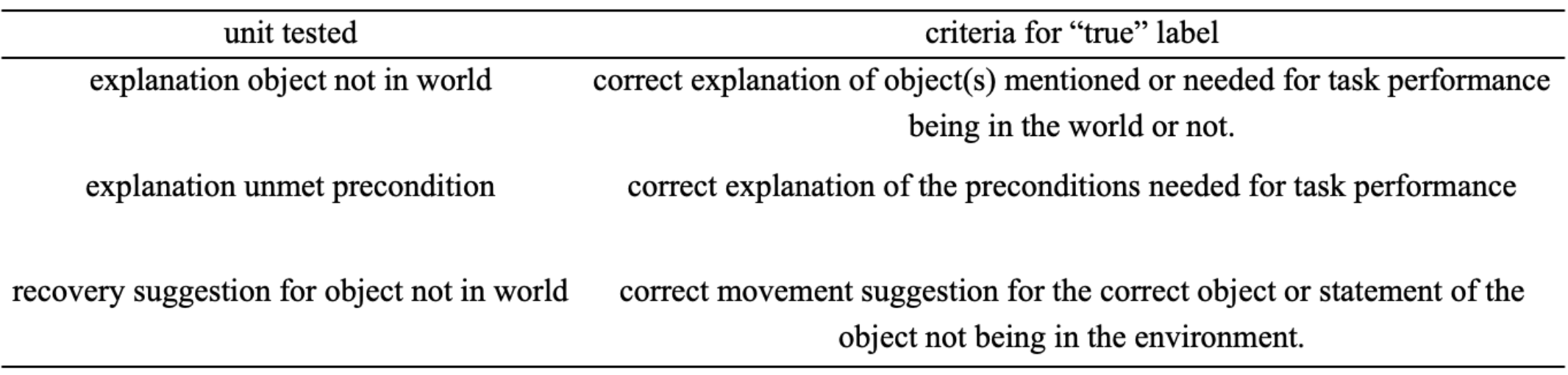}
	\caption{\label{fig:rating_criteria} Evaluation rating criteria}
\end{figure}

We assigned it a label of True if the method reply corresponded to the ground truth explanation in the dataset using the criteria in~\Cref{fig:rating_criteria}. 
We obtained two types of labels: 
(1) the label by one of the authors; 
and (2) the label by \textit{Mistral-Small-3.2-24B-Instruct-2506}\cite{mistral-small-3.2-24b-instruct-2506} as in \quotesasinlatex{LLM-as-a-judge}~\cite{gu2025surveyllmasajudge} (prompted with the label criteria displayed in \Cref{fig:rating_criteria}). 
The goal of using an LLM to label the data is to ensure unbiased labeling if only an author would label the data. 
In all figures in this subsection, we report accuracy as the \textit{average} between LLM and human labels,
where we report we obtained a satisfactory inter-rater reliability, i.e. \textit{Cohen's } $\kappa=0.91$.

\paragraph{Experiment B1: ablation of our framework}

\begin{figure}[t] 
	\centering
	\includegraphics[width=0.98\columnwidth]
    {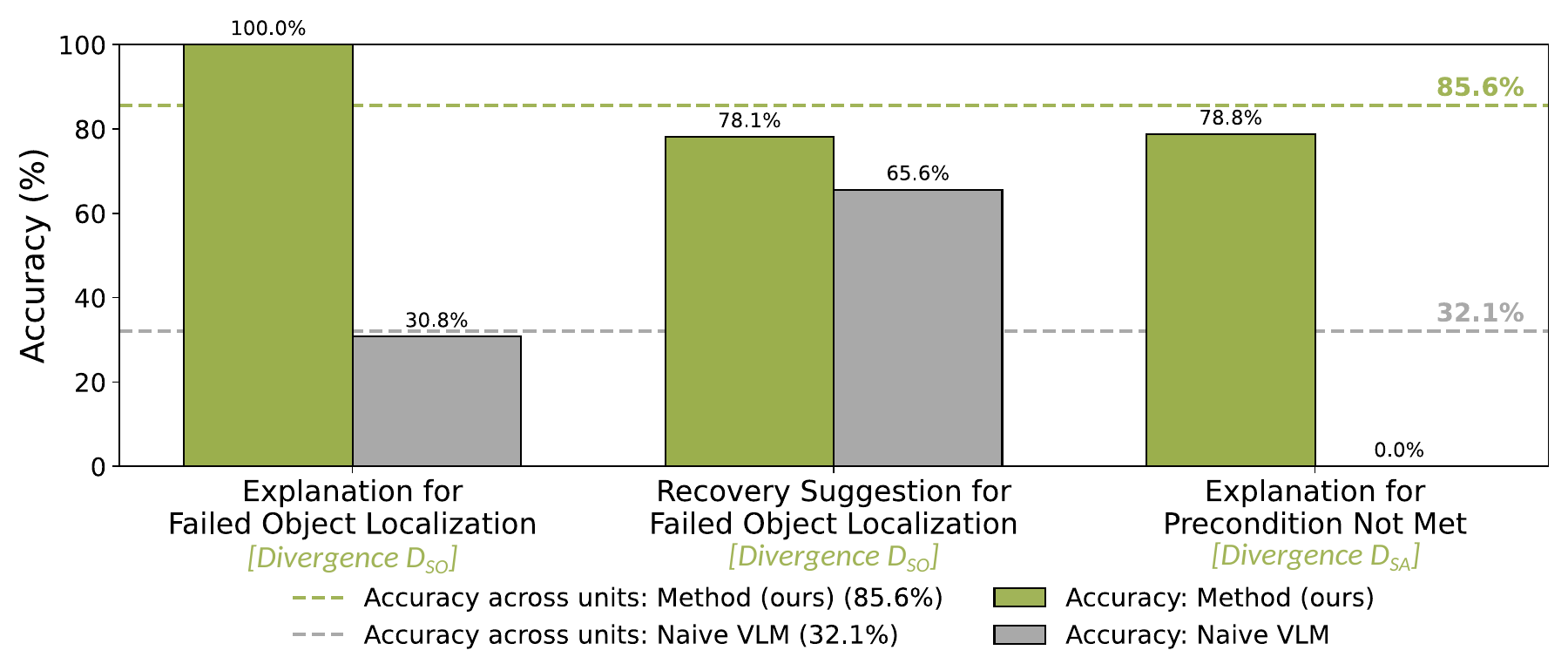}
	\caption{\label{fig:accuracy_unit} Accuracy scores for each unit for our method and the naive VLM}
\end{figure}

\begin{figure}[t] 
	\centering
	\includegraphics[width=0.95\columnwidth]{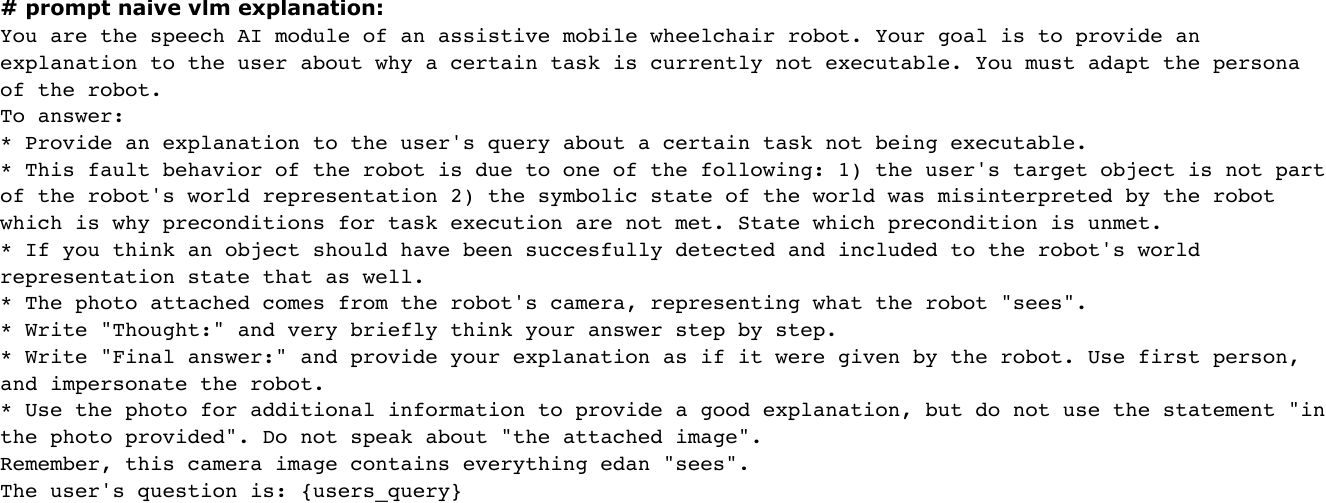}
	\caption{\label{fig:prompt_expl} Naive VLM prompt for explanations}
\end{figure}

\textit{\textbf{Question:} Are the explanation common sense, or is the robot's model required for correctly explaining and suggesting recovery strategies?}
\textbf{Benchmark with a Naive VLM:}
To draw a comparison, we ablated our framework by directly querying a \quotesasinlatex{\textit{naive}} VLM (again \textit{Mistral-Small-3.2}) with the same dataset including the image of the episodes. 
We used two prompts (one for the two explanation units in~\Cref{fig:prompt_expl} and one for the recovery suggestion unit), where general context about our robot, the task and the human's query was provided. 
However, the prompts did not contain information from robot's models. 
\textbf{Results:}
The accuracy was calculated for each unit separately, as shown in \Cref{fig:accuracy_unit}.
Examples of method and naive VLA replies are shown in~\Cref{fig:episode}.

\paragraph{Experiment B2: semantic representations}

\textit{\textbf{Question:} What is our reliance on semantically rich representations? }

By inspecting failed episodes of the unit for unmet preconditions, it seems that failures are mainly due to misinterpretations of uncommon English robot representations. For instance, the cabinet of the kitchen is represented on the robot's object database as \texttt{ikea\_bagganas}, the name of the product, which the LLM does not always identify as a cabinet, and sometimes mistakes it for an \quotesasinlatex{ikea bag}. 
We therefore repeat the unit test for \quotesasinlatex{Precondition not met} (\DivActIns), but augmented the relevant prompts with a \quotesasinlatex{translation dictionary} of uncommon terms. 

\begin{figure}[t] 
	\centering
	\includegraphics[width=0.6\columnwidth]{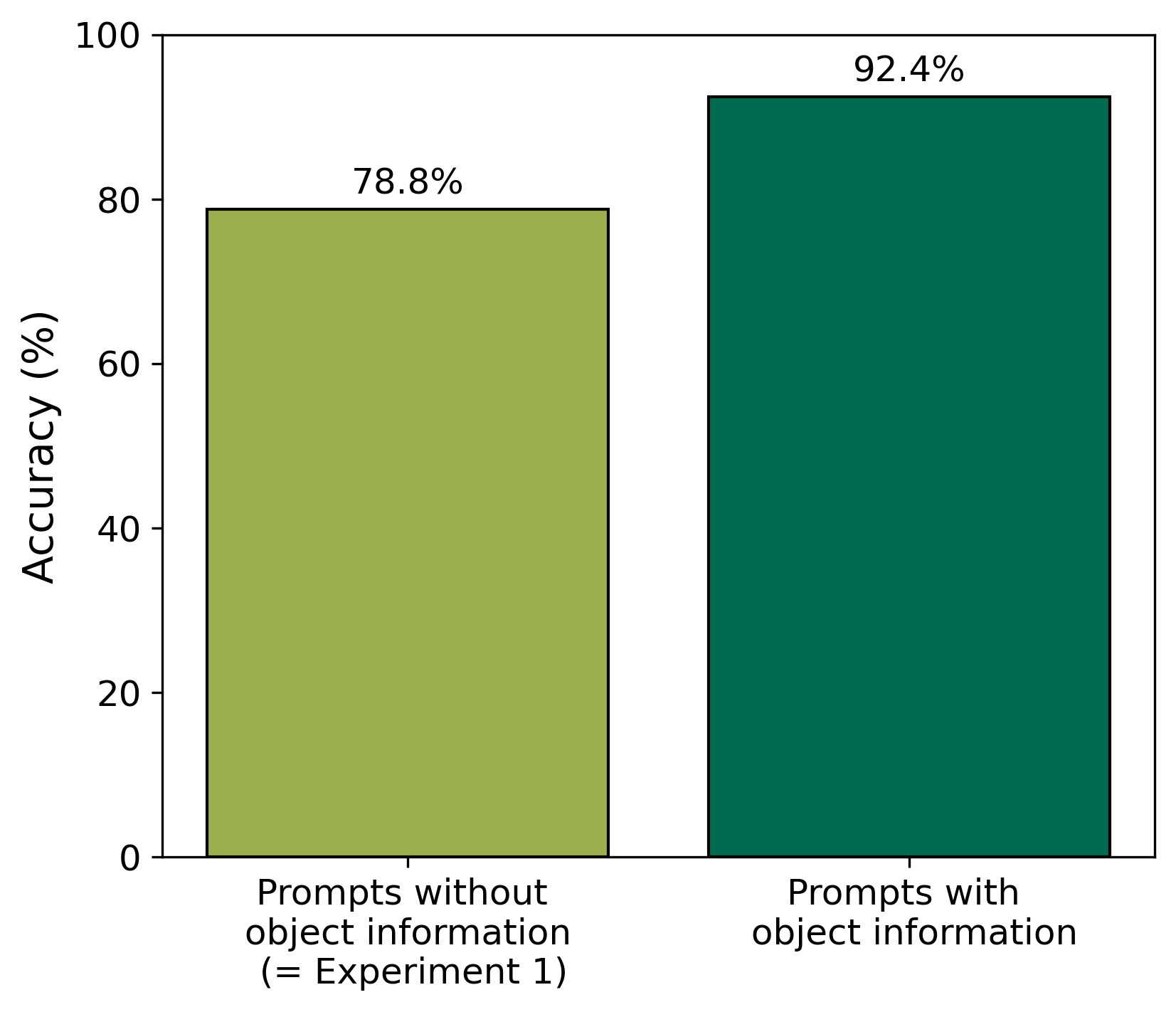}
	\caption{\label{fig:accuracy_object_info} Accuracy for the explanation for precondition not met unit for prompts without information about robotic object representation vs. prompts with information about it}
\end{figure}

\textbf{Results:}
As shown in \Cref{fig:accuracy_object_info}, providing a dictionary with translation of uncommon terms yielded an accuracy of 92.42\%, 13,63\% more than in Experiment 1.

\subsection{Discussion}
As shown in our robot experiments (\Cref{fig:results_expl,fig:results_recov}) and validated in our simulation unit tests (\Cref{fig:prompt_expl,fig:accuracy_object_info})  our method can create explanations and recovery suggestions for a variety of daily living objects and tasks situations, answering all real robot queries correctly on a long sequence and achieving between 78 and 100\% accuracy on the diverse unit tests. 
Experiment B1 shows that the model reconciliation outperforms a naive vision-only baseline, achieving 100\% accuracy in explaining object localization failures, 78.79\% in communicating symbolic state errors, and 78.12\% in recovery suggestions. 
Experiment B2 showed that adding a dictionary with definitions of uncommon terms (such as in the \texttt{ikea\_bagganas} kitchen cabinet example) further increases the accuracy from 78.8 to 92.4\% to the action reconciliation experiment, emphasizing the requirement of having clear robot representations that can be interpreted in English by the LLM. 
Furthermore, a naive VLM without the model reconciliation context has considerably lower accuracy in all conducted tests, which signifies that the LLM retrieves and provides the correct information from the modules of the robot to explain a failure situation, and a VLM by itself is not able to produce those explanations  by itself  

We higlight three limitations in our evaluation:  
(1) as shown in the unit test in~\Cref{fig:accuracy_object_info}, the recovery movement suggestions could improve, in the context of our robot. 
The accuracy drop seems to refer to incorrect platform suggestions by a VLM, which
could be due to limitations of VLMs with spatial reasoning\cite{pothiraj2025captureevaluatingspatialreasoning}, at least with the small model we used. 
(2) We noticed that (for small LLMs) adjectives can increase confusion. In one of our preliminary examples, the LLM would believe the \quotesasinlatex{greenish mug} would be in the world representation due to a green apple being present. 
(3) Finally, as our test cover only a limited number of situations of the daily living, our pilot study and unit tests will be extended to test the transparency and usability of the approach with
more diverse scenarios and objects.
To resolve these limitations a future user study would be appropriate, testing if the application of the method is indeed perceived as helpful and leads to more trust and improved human-robot-teaming. 

\section{Conclusion}
We introduced a model-reconciliation framework using foundation models to explain and recover from unexpected assistive robot behavior. 
The experiments show that the framework provides accurate language explanations and reasonable recovery suggestions, provided that the robot models contain representations that can be interpreted in English. 

In future work, we will explore explanations related to \textit{plan optimality} (explaining why a plan is optimal with longer planning horizons),
\textit{goals} (resolving mismatched goals between human and robot), and \textit{failed feasibility checks} (resolving situations where an action is possible in theory, but not in practice~\cite{bustamante2024feasibility}).
Furthermore, 
we will support multiple instances of the same object type, such as multiple drawers of a kitchen counter, which would require the human to specify object properties that disambiguate the query, such as \quotesasinlatex{why cannot I grasp the \textit{second drawer from the top}}.
Finally, 
we will consider comparisons with other user interface modalities, as visual explanations of a robotic wheelchair's model 
could lead to increased predictability and recovery times as well as a better understanding of the robot's inner workings \cite{RN13, RN14}.
\bibliographystyle{./style/IEEEtran}
\bibliography{bibliography}

@IEEEtranBSTCTL{IEEEexample:BSTcontrol,
	CTLuse_forced_etal       = "yes",
	CTLmax_names_forced_etal = "3",
    CTLuse_url = "no",
    CTLuse_doi = "no",
	CTLnames_show_etal       = "2" }

@misc{gu2025surveyllmasajudge,
      title={A Survey on LLM-as-a-Judge}, 
      author={Jiawei Gu and Xuhui Jiang and Zhichao Shi and Hexiang Tan and Xuehao Zhai and Chengjin Xu and Wei Li and Yinghan Shen and Shengjie Ma and Honghao Liu and Saizhuo Wang and Kun Zhang and Yuanzhuo Wang and Wen Gao and Lionel Ni and Jian Guo},
      year={2025},
      eprint={2411.15594},
      archivePrefix={arXiv},
      primaryClass={cs.CL},
      url={https://arxiv.org/abs/2411.15594}, 
}

@misc{pothiraj2025captureevaluatingspatialreasoning,
      title={CAPTURe: Evaluating Spatial Reasoning in Vision Language Models via Occluded Object Counting}, 
      author={Atin Pothiraj and Elias Stengel-Eskin and Jaemin Cho and Mohit Bansal},
      year={2025},
      eprint={2504.15485},
      archivePrefix={arXiv},
      primaryClass={cs.CV},
      url={https://arxiv.org/abs/2504.15485}, 
}

@misc{mistral-small-3.2-24b-instruct-2506,
  title        = {Mistral-Small-3.2-24B-Instruct-2506},
  author       = {{Mistral AI}},
  howpublished = {\url{https://huggingface.co/mistralai/Mistral-Small-3.2-24B-Instruct-2506}},
  note         = {A minor update over Small-3.1, improved instruction following, reduced repetition, and more robust function calling},
  month        = jul,
  year         = {2025},
  license      = {Apache-2.0}
}

@article{Hagengruber2025,
  title={An assistive robot that enables people with amyotrophia to perform sequences of everyday activities},
  author={Hagengruber, Annette and Quere, Gabriel and Iskandar, Maged and Bustamante, Samuel and Feng, Jianxiang and Leidner, Daniel and Albu-Sch{\"a}ffer, Alin and Stulp, Freek and Vogel, J{\"o}rn},
  journal={Scientific Reports},
  volume={15},
  number={1},
  pages={8426},
  year={2025},
  publisher={Nature Publishing Group UK London}
}

@inbook{NEURIPS2022_9d560961,
	author = {Wei, Jason and Wang, Xuezhi and Schuurmans, Dale and Bosma, Maarten and ichter, brian and Xia, Fei and Chi, Ed and Le, Quoc V and Zhou, Denny},
	booktitle = {Advances in Neural Information Processing Systems},
	editor = {S. Koyejo and S. Mohamed and A. Agarwal and D. Belgrave and K. Cho and A. Oh},
	pages = {24824-24837},
	publisher = {Curran Associates, Inc.},
	title = {Chain-of-Thought Prompting Elicits Reasoning in Large Language Models},
	url = {https://proceedings.neurips.cc/paper_files/paper/2022/file/9d5609613524ecf4f15af0f7b31abca4-Paper-Conference.pdf},
	volume = {35},
	year = {2022}
}

@misc{gao2021makingpretrainedlanguagemodels,
	title={Making Pre-trained Language Models Better Few-shot Learners}, 
	author={Tianyu Gao and Adam Fisch and Danqi Chen},
	year={2021},
	journal = {PsyArXiv},
}

@article{bustamante2024feasibility,
  author={Bustamante, Samuel and Rodríguez, Ismael and Quere, Gabriel and Lehner, Peter and Iskandar, Maged and Leidner, Daniel and Dömel, Andreas and Albu-Schäffer, Alin and Vogel, Jörn and Stulp, Freek},
  journal={IEEE Robotics and Automation Letters}, 
  title={Feasibility Checking and Constraint Refinement for Shared Control in Assistive Robotics}, 
  year={2024},
  volume={9},
  number={9},
  pages={8019-8026},
  doi={10.1109/LRA.2024.3430710}
}

@inproceedings{RN50,
	author = {Blankenburg, Janelle and Zagainova, Mariya and Simmons, S. Michael and Talavera, Gabrielle and Nicolescu, Monica and Feil-Seifer, David},
	title = {Human-robot collaboration and dialogue for fault recovery on hierarchical tasks},
	booktitle = {Social robotics: Proceedings of the 12th International Conference},
	editor = {Wagner, Alan R. and Feil-Seifer, David and Haring, Kerstin S. and Rossi, Silvia and Williams, Thomas and He, Hongsheng and Ge, Shuzhi Sam},
	location = {Golden, CO, USA},
	pages = {144-156},
	publisher = {Springer},
	year = {2020},
	doi = {10.1007/978-3-030-62056-1_13},
}

@inproceedings{RN14,
	author = {Brooks, Connor and Szafir, Daniel},
	title = {Visualization of intended assistance for acceptance of shared control},
	booktitle = {2020 IROS},
	publisher = {IEEE},
	pages = {11425-11430},
	ISBN = {1728162122},
	year = {2020},
	type = {Conference Proceedings},
	doi = {10.1109/IROS45743.2020.9340964}
}

@inproceedings{Bustamante2022cats,
	author = {Bustamante, Samuel and Quere, Gabriel and Leidner, Daniel and Vogel, Jörn and Stulp, Freek},
	title = {{CATs: Task planning for shared control of assistive robots with variable autonomy}},
	booktitle = {2022 ICRA},
	publisher = {IEEE},
	pages = {3775-3782},
	ISBN = {1728196817},
	doi = {10.1109/ICRA46639.2022.9811360},
	year = {2022},
	type = {Conference Proceedings}
}

@online{schluntz2024building,
  author       = {Erik Schluntz and Barry Zhang},
  title        = {Building Effective Agents},
  year         = {2024},
  url          = {https://www.anthropic.com/engineering/building-effective-agents},
  note         = {Engineering at Anthropic blog},
}

@inproceedings{RN133,
	author = {Chakraborti, Tathagata and Sreedharan, Sarath and Zhang, Yu and Kambhampati, Subbarao},
	title = {Plan explanations as model reconciliation: Moving beyond explanation as soliloquy},
	booktitle = {Proceedings of the 26th International Joint Conference on Artificial Intelligence (IJCAI)},
	pages = {156-163},
	year = {2017},
	doi = {10.24963/ijcai.2017/23},

}

@inproceedings{RN51,
	author = {Chance, Greg and Camilleri, Antonella and Winstone, Benjamin and Caleb-Solly, Praminda and Dogramadzi, Sanja},
	title = {An assistive robot to support dressing-strategies for planning and error handling},
	booktitle = {2016 6th IEEE International Conference on Biomedical Robotics and Biomechatronics (BioRob)},
	publisher = {IEEE},
	pages = {774-780},
	ISBN = {1509032878},
	year = {2016},
	type = {Conference Proceedings},
	doi = {10.1109/biorob.2016.7523721}
}

@article{RN80,
	author = {Eiband, Thomas and Willibald, Christoph and Tannert, Isabel and Weber, Bernhard and Lee, Dongheui},
	title = {Collaborative programming of robotic task decisions and recovery behaviors},
	journal = {Autonomous Robots},
	volume = {47},
	number = {2},
	pages = {229-247},
	ISSN = {0929-5593},
	year = {2023},
	type = {Journal Article},
	url = {https://doi.org/10.1007/s10514-022-10062-9}
}

@misc{RN192,
	author = {Ghallab, Malik   and Knoblock, Craig   and Wilkins, David   and Barrett, Anthony   and Christianson, David   and Friedman, Michael   and Kwok, Charles   and Golden, Keith   and Penberthy, Scott   and Smith, David   and Sun, Yixin   and Weld, Daniel},
	title = {{PDDL - The Planning Domain Definition Language}},
	url = {https://planning.wiki/_citedpapers/pddl1998.pdf},
	year = {1998},
	type = {Conference Paper}
}

@inproceedings{RN12,
	author = {Hori, Kazuki and Suzuki, Kanata and Ogata, Tetsuya},
	title = {Interactively Robot Action Planning with Uncertainty Analysis and Active Questioning by Large Language Model},
	booktitle = {2024 IEEE/SICE International Symposium on System Integration (SII)},
	publisher = {IEEE},
	pages = {85-91},
	ISBN = {9798350312072},
	year = {2024},
	type = {Conference Proceedings},

}

@inproceedings{RN82,
	author = {Klein, Stina and Huch, Jenny and Reißner, Nadine and Zwolsky, Pamina and Weitz, Katharina and Kraus, Matthias and André, Elisabeth},
	title = {Creating a Framework for a User-Friendly Cobot Failure Management in Human-Robot Collaboration},
	booktitle = {Companion of the 2024 ACM/IEEE International Conference on Human-Robot Interaction},
	pages = {618-622},
	year = {2024},
	type = {Conference Proceedings},
	url={http://doi.org/10.1145/3610978.3640591}
}

@book{RN173,
	author = {Leidner, Daniel Sebastian},
	title = {Cognitive reasoning for compliant robot manipulation},
	publisher = {Springer},
	volume = {23},
	year = {2019},
	type = {Book}
}

@article{RN1,
	author = {Miller, Tim},
	title = {Explanation in artificial intelligence: Insights from the social sciences},
	journal = {Artificial intelligence},
	volume = {267},
	pages = {1-38},
	ISSN = {0004-3702},
	year = {2019},
	type = {Journal Article},
	url = {https://doi.org/10.1016/j.artint.2018.07.007}
}

@inproceedings{RN191,
	author = {Quere, Gabriel and Hagengruber, Annette and Iskandar, Maged and Bustamante, Samuel and Leidner, Daniel and Stulp, Freek and Vogel, Jörn},
	title = {Shared control templates for assistive robotics},
	booktitle = {2020 ICRA},
	publisher = {IEEE},
	pages = {1956-1962},
	year = {2020},
	type = {Conference Proceedings}
}

@article{RN174,
	author = {Radvansky, Gabriel A and Zacks, Jeffrey M},
	title = {Event perception},
	journal = {Wiley Interdisciplinary Reviews: Cognitive Science},
	volume = {2},
	number = {6},
	pages = {608-620},
	year = {2011},
	type = {Journal Article}
}

@article{RN146,
	author = {Sarch, Gabriel and Wu, Yue and Tarr, Michael J and Fragkiadaki, Katerina},
	title = {Open-ended instructable embodied agents with memory-augmented large language models},
	journal = {PsyArXiv},
	year = {2023},
	type = {Journal Article},
	url = {https://doi.org/10.48550/arXiv.2310.15127}
}

@article{RN33,
	author = {Tabrez, Aaquib and Luebbers, Matthew B and Hayes, Bradley},
	title = {A survey of mental modeling techniques in human-robot teaming},
	journal = {Current Robotics Reports},
	volume = {1},
	pages = {259-267},
	year = {2020},
	type = {Journal Article},
	url = {https://doi.org/10.1007/s43154-020-00019-0}
}

@inproceedings{RN8,
	author = {Vogel, Jörn and Hagengruber, Annette and Iskandar, Maged and Quere, Gabriel and Leipscher, Ulrike and Bustamante, Samuel and Dietrich, Alexander and Höppner, Hannes and Leidner, Daniel and Albu-Schäffer, Alin},
	title = {{EDAN: An EMG-controlled daily assistant to help people with physical disabilities}},
	booktitle = {2020 IEEE/RSJ International Conference on Intelligent Robots and Systems (IROS)},
	publisher = {IEEE},
	pages = {4183-4190},
	ISBN = {1728162122},
	doi = {10.1109/IROS45743.2020.9341156},
	year = {2020},
	type = {Conference Proceedings}
}

@article{RN57,
	author = {Wallkötter, Sebastian and Tulli, Silvia and Castellano, Ginevra and Paiva, Ana and Chetouani, Mohamed},
	title = {{Explainable Embodied Agents Through Social Cues: A Review}},
	journal = {ACM Transactions on Human-Robot Interaction (THRI)},
	volume = {10},
	number = {3},
	pages = {1-24},
	ISSN = {2573-9522},
	year = {2021},
	type = {Journal Article},
	url ={https://doi.org/10.1145/3457188}
}

@article{RN190,
	author = {Wimmer, Heinz and Perner, Josef},
	title = {Beliefs about beliefs: Representation and constraining function of wrong beliefs in young children's understanding of deception},
	journal = {Cognition},
	volume = {13},
	number = {1},
	pages = {103-128},
	ISSN = {0010-0277},
	url = {https://doi.org/10.1016/0010-0277(83)90004-5},
	year = {1983},
	type = {Journal Article}
}

@inproceedings{RN52,
	author = {Wu, Ruichao and Kortik, Sitar and Santos, Christoph Hellmann},
	title = {Automated behavior tree error recovery framework for robotic systems},
	booktitle = {2021 ICRA},
	publisher = {IEEE},
	pages = {6898-6904},
	ISBN = {1728190770},
	year = {2021},
	type = {Conference Proceedings},
	url = {https://doi.org/10.1109/icra48506.2021.9561002 }
}

@techreport{Behery2016,
	type = {{Master's} {Thesis}},
	title = {A Knowledge-Based Activity Representation for Shared Autonomy Teleoperation of Robotic Arms},
	institution = {RWTH-Aachen University},
	author = {M. {Behery}},
	year = {2016},
}

@inproceedings{moorman2025bidirectional,
      title={Bi-Directional Mental Model Reconciliation for Human-Robot Interaction with Large Language Models}, 
      author={Nina Moorman and Michelle Zhao and Matthew B. Luebbers and Sanne Van Waveren and Reid Simmons and Henny Admoni and Sonia Chernova and Matthew Gombolay},
      year={2025},
	   booktitle = {Workshop in Advancing Artificial Intelligence through Theory of Mind, AAAI},
      eprint={2503.07547},
      archivePrefix={arXiv},
      primaryClass={cs.RO},
}

@inproceedings{vasileiou2023,
      title={On Exploiting Hitting Sets for Model Reconciliation}, 
      author={Stylianos Loukas Vasileiou and Alessandro Previti and William Yeoh},
      year={2023},
      booktitle={AAAI Conference on Artificial Intelligence (AAAI-21)},
      eprint={2012.09274},
      archivePrefix={arXiv},
      primaryClass={cs.AI},
}

@article{Dung2022,
author = {Dung, Ho and Son, Tran},
year = {2022},
month = {08},
pages = {27-48},
title = {On Model Reconciliation: How to Reconcile When Robot Does not Know Human's Model?},
volume = {364},
journal = {Electronic Proceedings in Theoretical Computer Science},
}

@article{Sakagami2023,
	month = {November},
	title = {Robotic world models {--} conceptualization, review, and engineering best practices},
	editor = {Sariyildiz, Emre},
	year = {2023},
	volume = {10},
	publisher = {Frontiers Media S.A},
	author = {Sakagami, Ryo and Lay, Florian Samuel and D{\"o}mel, Andreas and Schuster, Martin and Albu-Sch{\"a}ffer, Alin Olimpiu and Stulp, Freek},
	journal = {Frontiers in Robotics and AI},
	url = {https://elib.dlr.de/198741/},

}

@inproceedings{RN39,
	author = {Zakershahrak, Mehrdad and Marpally, Shashank Rao and Sharma, Akshay and Gong, Ze and Zhang, Yu},
	title = {Order matters: Generating progressive explanations for planning tasks in human-robot teaming},
	booktitle = {2021 ICRA},
	publisher = {IEEE},
	pages = {3751-3757},
	ISBN = {1728190770},
	year = {2021},
	type = {Conference Proceedings},
	url = {https://doi.org/10.1109/icra48506.2021.9561762}
}

@article{RN148,
	author = {Zha, Lihan and Cui, Yuchen and Lin, Li-Heng and Kwon, Minae and Arenas, Montserrat Gonzalez and Zeng, Andy and Xia, Fei and Sadigh, Dorsa},
	title = {Distilling and retrieving generalizable knowledge for robot manipulation via language corrections},
	journal = {PsyArXiv},
	year = {2023},
	type = {Journal Article},
}

@inproceedings{RN13,
	author = {Zolotas, Mark and Demiris, Yiannis},
	title = {Transparent intent for explainable shared control in assistive robotics},
	booktitle = {Proceedings of the Twenty-Ninth International Conference on International Joint Conferences on Artificial Intelligence},
	pages = {5184-5185},
	year = {2021},
	type = {Conference Proceedings}
}


%
%
%
%
%
%

\end{document}